\documentclass[10pt,twocolumn,letterpaper]{article}

\usepackage{cvpr}      

\usepackage{booktabs}
\usepackage{array}

\definecolor{cvprblue}{rgb}{0.21,0.49,0.74}
\usepackage[pagebackref,breaklinks,colorlinks,allcolors=cvprblue]{hyperref}

\def\paperID{6} 
\def\confName{CVPR}
\def\confYear{2026}

\title{PGcGAN: Pathological Gait-Conditioned GAN for Human Gait Synthesis}

\author{Mritula Chandrasekaran \and Jarek Francik \and Dimitrios Makris\\
Kingston University\\
Kingston Upon Thames, London, UK\\
{\tt\small c.mritula@kingston.ac.uk \quad jarek@kingston.ac.uk \quad d.makris@kingston.ac.uk }
\and
Sanket Kachole\\
Indiana University Health \\
Indianapolis, Indiana\\
{\tt\small skachole@iu.edu}
}

\begin{document}
\maketitle
\begin{abstract}
Pathological gait analysis is constrained by limited and variable clinical datasets, which restrict the modeling of diverse gait impairments. To address this challenge, we propose a Pathological Gait-conditioned Generative Adversarial Network (PGcGAN) that synthesises pathology-specific gait sequences directly from observed 3D pose keypoint trajectories data. The framework incorporates one-hot encoded pathology labels within both the generator and discriminator, enabling controlled synthesis across six gait categories. The generator adopts a conditional autoencoder architecture trained with adversarial and reconstruction objectives to preserve structural and temporal gait characteristics. Experiments on the Pathological Gait Dataset demonstrate strong alignment between real and synthetic sequences through PCA and t-SNE analyses, visual kinematic inspection, and downstream classification tasks. Augmenting real data with synthetic sequences improved pathological gait recognition across GRU, LSTM, and CNN models, indicating that pathology-conditioned gait synthesis can effectively support data augmentation in pathological gait analysis.
\end{abstract}    
\section{Introduction}
\label{sec:intro}

Our work introduces a data-driven framework that learns gait variability directly from observed motion data, enabling the modelling of complex spatiotemporal dynamics and condition-specific gait patterns. Rather than replacing biomechanical understanding, the approach complements it by leveraging generative modelling to capture variability that may be difficult to represent through analytical models. A conditional Generative Adversarial Network (cGAN) is developed for scalable synthesis of diverse gait sequences. The model generates pathology-specific gait patterns while preserving biomechanical structure and inter-subject variability.

The proposed framework is evaluated on the Pathological Gait Dataset \cite{jun2020pathological}. Structural consistency between real and synthetic sequences is assessed. 
Additionally, downstream classification experiments examine the utility of the generated data for gait recognition tasks. The results indicate that the synthesised sequences maintain biomechanically plausible motion patterns and provide useful augmentation for computer vision–based pathological gait analysis.

\section{Literature Review}
\label{literaturereview}
Generative modeling has become an increasingly important tool in gait analysis, particularly for addressing limitations related to dataset size, subject diversity, and annotation cost. By learning the statistical structure of motion data, generative models can synthesize new gait sequences that augment existing datasets and support robust machine learning models.

Early work on generative gait modeling focused primarily on adversarial learning frameworks. One of the most influential approaches is \textit{GaitGAN}, demonstrated that GANs can perform cross-view gait image synthesis while preserving identity information \cite{yu2017gaitgan}. 
Building on this idea, later methods explored the use of pose-guided and silhouette-based generation strategies. These approaches incorporated structural motion information into the generation process by conditioning synthesis on silhouettes \cite{Fan2020GaitPart, nie2021view} or joint trajectories \cite{min2024gaitma}. Such methods improved representation learning and visual fidelity, but their primary objective remained recognition performance rather than the synthesis of biomechanically meaningful gait motion.

More recent work has explored conditional and disentangled representation learning to improve controllability in generative gait models. Conditional GANs extend the framework by incorporating auxiliary variables into both the generator and discriminator, allowing the synthesis of samples conditioned on attributes such as pose, viewpoint, or subject identity \cite{bicer2024generative}. This mechanism enables targeted generation and improved representation learning. In parallel, adversarial autoencoders (AAEs) have been proposed to organise latent representations by aligning the encoded distribution with a predefined prior using adversarial training \cite{makhzani2015adversarial}. Such structured latent spaces facilitate the manipulation of gait attributes and support controlled synthesis of motion patterns. Despite these advances, most conditional frameworks focus on external covariates rather than biomechanical or pathological factors. In addition, evaluation often relies on visual inspection or recognition accuracy without explicitly assessing the structural or dynamic plausibility of generated motion \cite{guo2024online, barsoum2018hp, radford2015unsupervised, li2025morph}. Burges et al. recently demonstrated that conditioning mechanisms can improve identity consistency in gait synthesis, although their work focused primarily on standard gait patterns rather than pathological variations \cite{burges2024gait}.

In general, the generative modeling of pathological gait remains relatively underexplored. Pathological gait datasets are typically small and difficult to collect, creating a strong motivation for synthetic data generation. Several studies investigated the use of GANs for generating abnormal gait sequences. For example, Wang et al. proposed a GAN-based framework to augment impaired gait datasets for classification tasks \cite{wang2021human}. Similarly, Sadeghzadehyazdi et al. modeled Parkinsonian gait using adversarial training on spatiotemporal gait features \cite{sadeghzadehyazdi2021modeling}. Song et al. introduced a GAN-based sequence generation method for abnormal gait recognition \cite{song2020novel}, while FoGGAN focused on synthesizing freezing-of-gait episodes to improve detection robustness \cite{peppes2023foggan}. Although these approaches demonstrate the potential of generative models for pathological gait analysis, they are typically limited to specific disorders or controlled experimental conditions.

Overall, prior work shows that GAN-based models can successfully learn gait distributions and support data augmentation for recognition tasks. However, most existing approaches lack explicit conditioning on pathological gait categories and rarely incorporate structural validation of generated motion. These limitations motivate the development of generative frameworks that explicitly encode pathology information while preserving the biomechanical structure of gait sequences.

\section{Methodology}
\label{sec:methodology}

\begin{figure*}[t]
\centering
\includegraphics[width=0.6\linewidth]{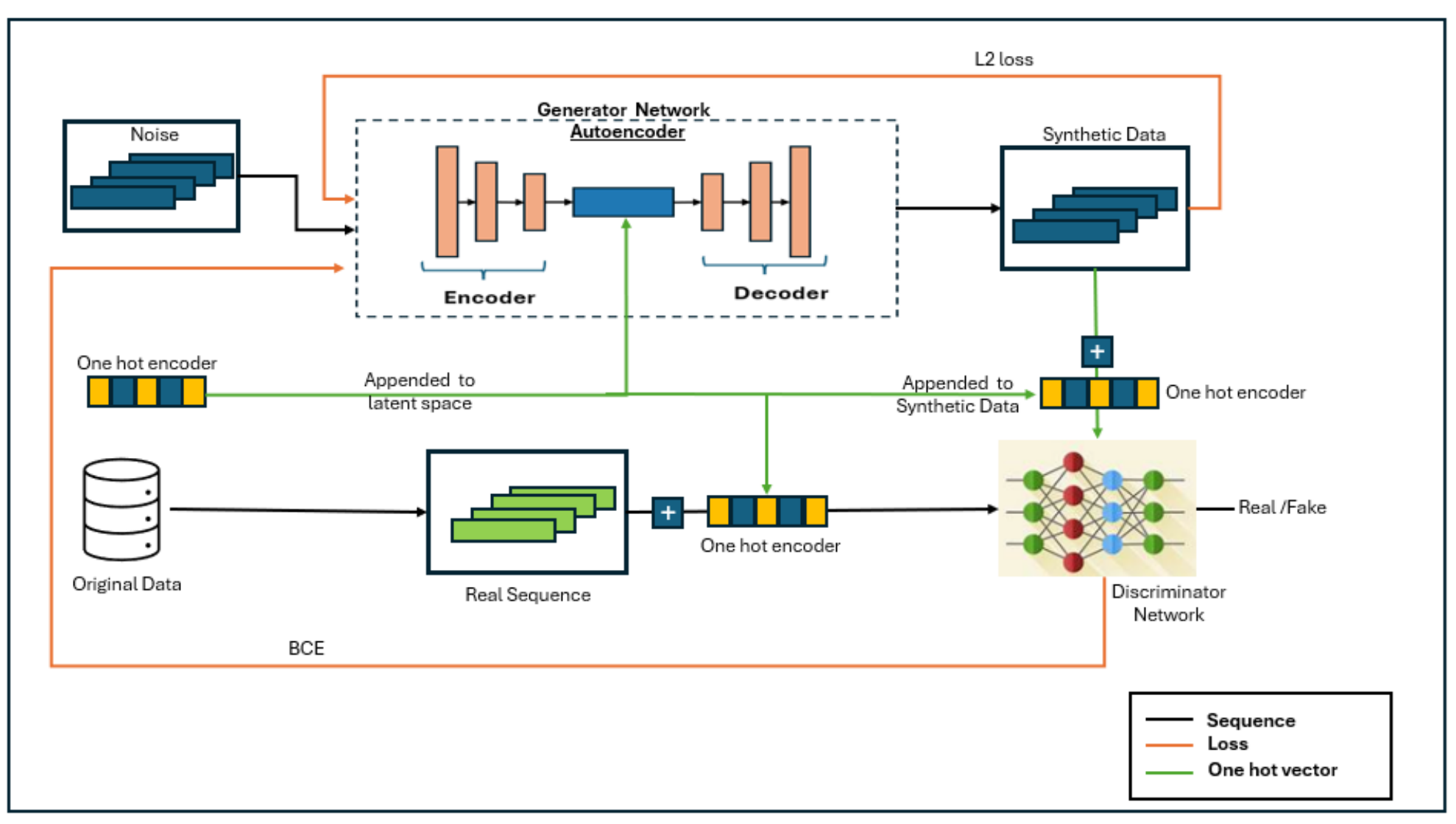}
\caption{Overview of the proposed Pathological Gait Conditioned GAN (PGcGAN) architecture. 
Gaussian noise and one-hot encoded pathology labels are used to condition the generator, 
which produces synthetic gait sequences through an encoder–decoder structure. 
Both real and generated sequences are combined with pathology labels and evaluated by the 
discriminator to distinguish real from synthetic motion patterns.}
\label{fig:methodology}
\end{figure*}

This section describes the proposed \textit{Pathological Gait Conditioned GAN (PGcGAN)}, a conditional generative framework designed to synthesize pathology-specific gait sequences. While inspired by \cite{bicer2024generative}, the architecture is extended through explicit pathology conditioning. The model combines adversarial learning with conditional representation learning to generate diverse gait trajectories while preserving biomechanical structure and temporal coherence. The overall architecture consists of two main components: a conditional generator and a conditional discriminator. Pathology labels are incorporated into both networks to guide the generation process and enforce class consistency.

\subsection{Problem Formulation}

Let a gait sequence be represented as
\begin{equation}
X = \{x_1, x_2, ..., x_T\}, \quad x_t \in \mathbb{R}^{d}
\end{equation}

\noindent where $x_t$ denotes the joint feature vector at time step $t$, $T$ is the sequence length, and $d$ represents the dimensionality of the gait feature space. Each sequence is associated with a pathology label $y \in \{1,...,C\}$, where $C$ denotes the number of gait categories.

The objective of the generative model is to learn a conditional distribution $p(X|y)$ that allows the generation of realistic gait sequences consistent with the specified pathology condition.

\subsection{Conditional Generator}

The generator $G$ is implemented as a conditional autoencoder that transforms a stochastic input into a temporally structured gait sequence. A Gaussian noise tensor $n \sim \mathcal{N}(0,I)$ with dimensionality matching the gait representation is first mapped into a latent representation through an encoder network $z = E(n)$.

To incorporate pathology information, the latent representation is concatenated with a one-hot encoded pathology label $y$, forming a conditioned latent vector $z' = [z; y]$.

This conditioned representation is then passed to the decoder to produce a synthetic gait sequence $\hat{X} = D(z')$.

Both the encoder and decoder are implemented using temporal convolutional blocks with ReLU activations, enabling the model to capture short-term motion patterns and temporal dependencies within gait sequences. The reconstruction pathway further encourages the generator to preserve structural properties of real gait data.

\subsection{Conditional Discriminator}

The discriminator $D$ is trained to distinguish real gait sequences from generated ones while considering the associated pathology condition. Real and synthetic sequences are concatenated with their corresponding one-hot encoded labels before being passed to the discriminator.

\begin{equation}
D([X; y]) \rightarrow 1, \quad D([\hat{X}; y]) \rightarrow 0
\end{equation}

The discriminator consists of stacked temporal convolutional layers followed by fully connected layers and a sigmoid output unit. Early convolutional layers capture local motion dynamics, while deeper layers aggregate longer temporal patterns that characterize different gait conditions. Spectral normalization is applied to convolutional layers to improve training stability.

\subsection{Training Objective}

The model is trained using a combination of adversarial and reconstruction objectives. The adversarial loss encourages the generator to produce sequences that are indistinguishable from real gait data

\begin{equation}
\begin{aligned}
\mathcal{L}_{adv} &= 
\mathbb{E}_{X,y \sim p_{data}}[\log D(X,y)] \\
&\quad + 
\mathbb{E}_{n,y \sim p_z}[\log(1 - D(G(n,y),y))]
\end{aligned}
\end{equation}

To ensure structural consistency between generated and real sequences, an $\mathcal{L}_2$ reconstruction loss is applied

\begin{equation}
\mathcal{L}_{rec} = ||X - \hat{X}||_2^2
\end{equation}

The final generator objective combines these terms

\begin{equation}
\mathcal{L}_{gen} = \lambda_{adv}\mathcal{L}_{adv} + \lambda_{rec}\mathcal{L}_{rec}
\end{equation}

\noindent where $\lambda_{adv}$ and $\lambda_{rec}$ control the balance between realism and structural fidelity.

\subsection{Training Procedure}

The generator and discriminator are optimized jointly in an adversarial training scheme. During each training iteration, a batch of real gait sequences and corresponding pathology labels is sampled from the dataset. Synthetic sequences are generated using sampled noise vectors conditioned on the same labels. The discriminator is first updated to correctly classify real and synthetic sequences, after which the generator is updated to minimize the combined adversarial and reconstruction objectives.

Training proceeds until the discriminator can no longer reliably distinguish between real and synthetic gait sequences under pathology conditioning, indicating that the generator has learned a realistic distribution of gait patterns.

\section{Experimental Evaluation}
This section presents the experimental evaluation of the proposed PGcGAN using the Pathological Gait Dataset \cite{jun2020pathological}. Structural analysis and classification experiments are conducted to assess the realism of the generated gait sequences and their effectiveness for pathological gait recognition.

\subsection{tSNE Analysis of Joint Angle Trajectories}
\begin{figure*}[!t]
    \centering
    \begin{subfigure}[t]{0.30\textwidth}
        \centering
        \includegraphics[width=\linewidth]{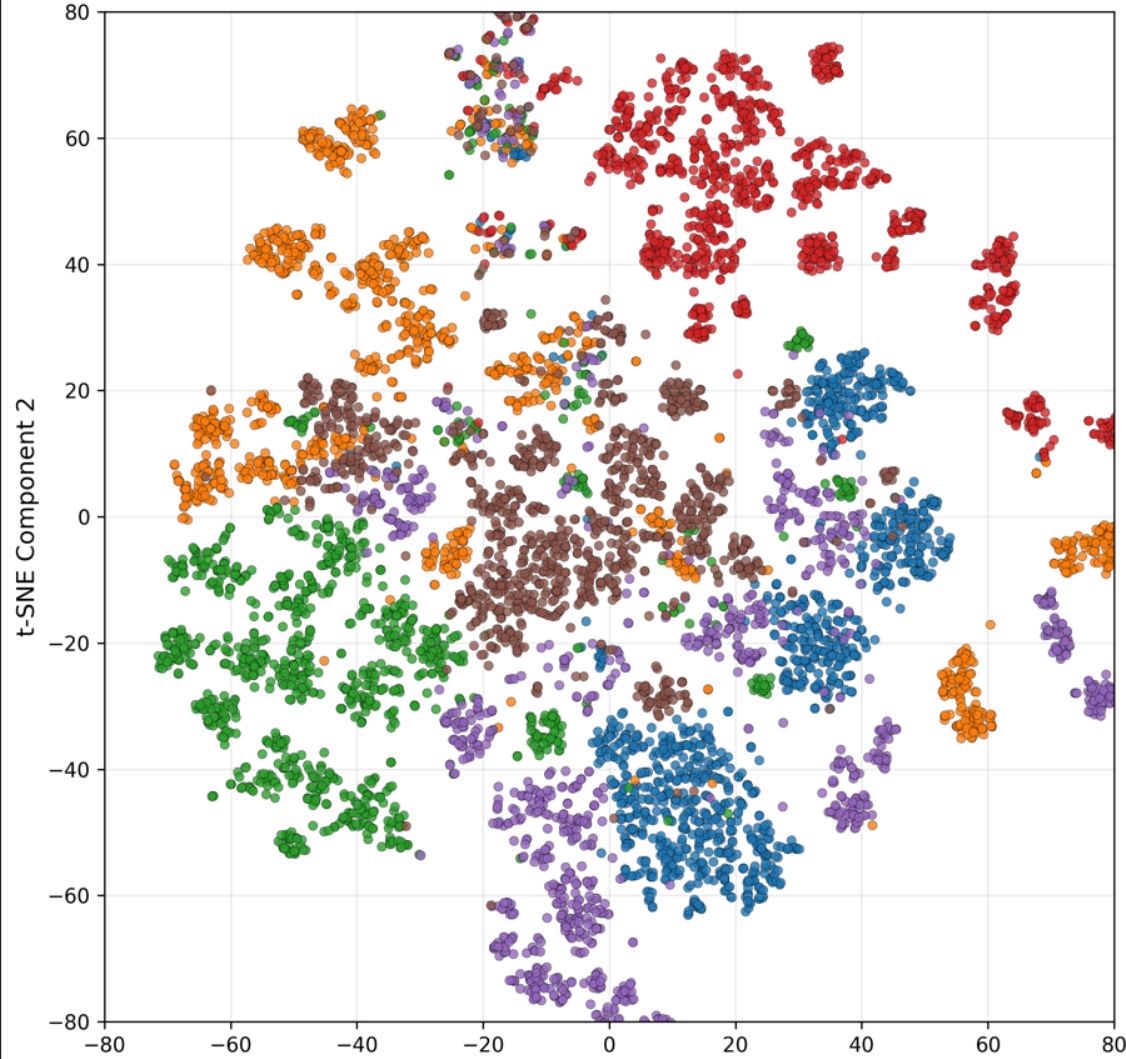}
        \caption{Real gait features}
        \label{fig:tsne_real}
    \end{subfigure}
    \hspace{0.02\textwidth}
    \begin{subfigure}[t]{0.45\textwidth}
        \centering
        \includegraphics[width=\linewidth]{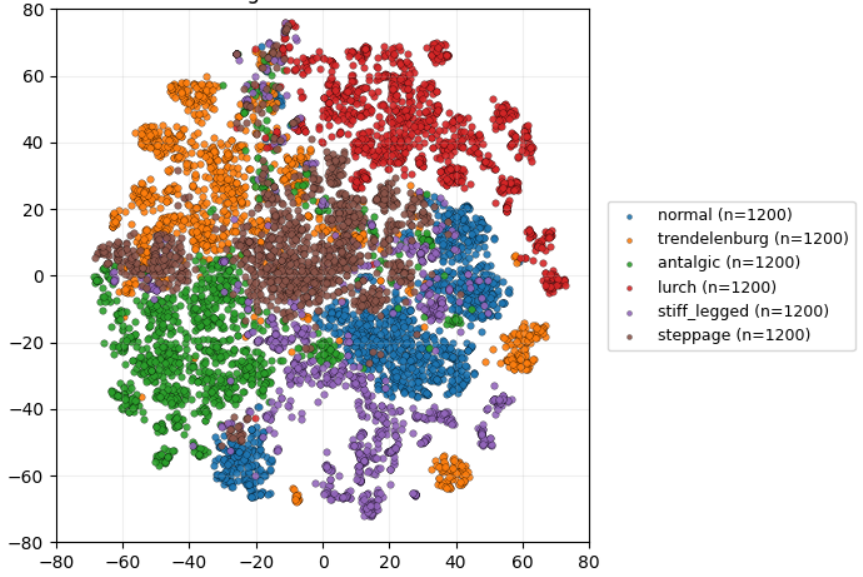}
        \caption{Synthetic gait features}
        \label{fig:tsne_gan}
    \end{subfigure}
    \caption{t-SNE visualization comparing the distribution of real and PGcGAN-generated gait features.}    \label{fig:tsne_comparison_realvsgan}
\end{figure*}

Following the evaluation strategy in \cite{yoon2019time}, we apply t-SNE to compare real and GAN-generated gait features. The t-SNE embedding in Figure~\ref{fig:tsne_comparison_realvsgan} reveals strong alignment between real and synthetic gait samples, suggesting that the proposed model preserves the latent structure and variability of pathological gait patterns.

\subsection{Classifier Performance}
\label{ClassifierPerformance}
Three deep learning models, GRU, LSTM, and CNN, were trained under three configurations: (a) only real data, (b) only GAN-generated synthetic data, and (c) a combination of real and GAN-generated synthetic data. To examine the effect of sequence filtering, sequences with a length greater than or equal to 60 frames were considered, following the protocol used in the dataset study \cite{jun2020pathological}. This resulted in a filtered dataset of 7,157 sequences after removing shorter samples. The classification results are presented in Table \ref{tab:classification_results_7157}. 

 The results in Table~\ref{tab:classification_results_7157} highlight two key findings.  First, the models trained on synthetic-only data consistently underperformed compared to real-only training, reflecting the limitations of GAN-generated sequences in capturing the full variability of natural gait. Nonetheless, synthetic-only accuracies remained competitive and in several cases outperformed the benchmark results reported by Jun et al. \cite{jun2020pathological}, suggesting that GAN data retains discriminative value.  Second, the inclusion of synthetic data alongside real data consistently improved classifier performance across most models. GRU improved from 91.87\% to 92.61\% and CNN from 87.90\% to 89.56\%, while LSTM maintained comparable accuracy.



\begin{table}[t]
\centering
\small
\resizebox{\columnwidth}{!}{%
\begin{tabular}{lccc}
\toprule
\textbf{Model} & \textbf{Real} & \textbf{Synthetic} & \textbf{Real + Synthetic} \\
\midrule
GRU  & 91.87 & 87.65 & \textbf{92.61} \\
LSTM & \textbf{90.48} & 85.26 & 90.08 \\
CNN  & 87.90 & 83.71 & \textbf{89.56} \\
\bottomrule
\end{tabular}%
}
\caption{Classification accuracy (\%) of GRU, LSTM, and CNN on the filtered dataset (7157 sequences, $>$60 frames). Synthetic-only training performs worse than real-only, while augmentation improves performance.}
\label{tab:classification_results_7157}
\end{table}


\subsection{Comparison with Existing Synthetic Gait Generation Approaches}
\label{ComparisonwithExistingSyntheticGaitGenerationApproaches}

Table~\ref{tab:comparison_literature} compares the proposed approach with a previous existing synthetic gait generation method, which focused on normal gait and evaluated realism using trajectory similarity or distributional alignment. In contrast, our method synthesizes multiple pathological gait types and validates them using both structural similarity measures and visual kinematic analysis, and still achieves a high value of$R^2$. In addition, Table~\ref{tab:comparison_jun2020} shows that our approach outperforms previous work \cite{jun2020pathological}, confirming the advantage of augmenting real data with the generated sequences. These results demonstrate that the proposed framework effectively preserves gait structure while enhancing pathological gait recognition.

\begin{table}[t]
\centering
\small
\resizebox{\columnwidth}{!}{%
\begin{tabular}{p{2.8cm} p{3.9cm} c}
\toprule
\textbf{Study} & \textbf{Validation Approach} & \textbf{Mean $R^2$} \\
\midrule
Kim \& Hargrove \cite{kim2023synthetic} &
Trajectory similarity and prosthetic control performance metrics &
0.97 \\
\textbf{PGcGAN \newline (Our Study)} &
Trajectory similarity metrics + visual kinematic envelope comparison &
0.94 \\
\bottomrule
\end{tabular}%
}
\caption{Comparison of synthetic gait validation results for normal gait trajectories with existing literature.}
\label{tab:comparison_literature}
\end{table}

\begin{table}[t]
\centering
\small
\setlength{\tabcolsep}{4pt}
\begin{tabular}{p{3cm} c p{2.5cm}}
\toprule
\textbf{Method} & \textbf{Accuracy (\%)} & \textbf{Dataset Used} \\
\midrule
Jun et al. (2020) \cite{jun2020pathological} & 90.13 & Real Only \\
PGcGAN \newline Ours (GRU) & \textbf{92.61}  & Real + Synthetic \\
\bottomrule
\end{tabular}
\caption{Comparison with the previously reported baseline using the same pathological gait dataset \cite{jun2020pathological}.}
\label{tab:comparison_jun2020}
\end{table}

\subsection{Conclusion}
Our work presented  Pathological Gait-Conditioned GAN (PGcGAN), a conditional GAN framework tailored for pathological gait synthesis,
with the novel contribution of incorporating explicit pathology conditioning through one-hot
labels injected at multiple stages of both generator and discriminator pathways. This design
enabled the generation of diverse, pathology-specific gait sequences while preserving identity
and temporal structure, positioning GAN-based synthesis as a scalable complement to physics-
based simulation.


{
    \small
    \bibliographystyle{ieeenat_fullname}
    \bibliography{main}
}


\end{document}